\documentclass[letterpaper,10pt,conference]{ieeeconf}  % Comment this line out if you need a4paper

\IEEEoverridecommandlockouts                              % This command is only needed if 
\usepackage{mymacros}
\usepackage{times} % assumes new font selection scheme installed
\usepackage{graphicx}

\usepackage[T1]{fontenc}
\usepackage{color}
\usepackage{hyperref}
\usepackage{pifont}% http://ctan.org/pkg/pifont

\let\labelindent\relax
\usepackage{enumitem}

\usepackage{subcaption}

\title{\LARGE \bf
Prototypical Contrastive Transfer Learning\\for Multimodal Language Understanding
}

\author{
    Seitaro Otsuki$^{1}$, Shintaro Ishikawa$^{1}$ and Komei Sugiura$^{1}$% <-this % stops a space
    \thanks{$^{1}$The authors are with Keio University, 3-14-1 Hiyoshi, Kohoku, Yokohama, Kanagawa 223-8522, Japan.
         {\tt\small \{otsu8sei14, shin.0116, komei.sugiura\}@keio.jp}
    }%
}

\begin{document}

\maketitle
\thispagestyle{empty}
\pagestyle{empty}

\begin{abstract}
Although domestic service robots are expected to assist individuals who require support, they cannot currently interact smoothly with people through natural language. For example, given the instruction ``Bring me a bottle from the kitchen,'' it is difficult for such robots to specify the bottle in an indoor environment.
Most conventional models have been trained on real-world datasets that are labor-intensive to collect, and they have not fully leveraged simulation data through a transfer learning framework.
In this study, we propose a novel transfer learning approach for multimodal language understanding called Prototypical Contrastive Transfer Learning (PCTL), which uses a new contrastive loss called Dual ProtoNCE. We introduce PCTL to the task of identifying target objects in domestic environments according to free-form natural language instructions.
To validate PCTL, we built new real-world and simulation datasets.
Our experiment demonstrated that PCTL outperformed existing methods. Specifically, PCTL achieved an accuracy of 78.1\%, whereas simple fine-tuning achieved an accuracy of 73.4\%.
\end{abstract}

% \vspace{-1mm}
\section{Introduction
\label{intro}
}
% \vspace{-1mm}
In our aging society, the demand for daily care and support is increasing, which is leading to a shortage of home care workers. Domestic service robots (DSRs) are gaining popularity as a solution because of their ability to physically assist individuals. However, DSRs currently lack the capability to smoothly interact with people through natural language. To train their language comprehension models, it is desirable to use data collected in real-world environments. However, collecting and annotating such real-world data can be labor-intensive. By contrast, collecting training data using a simulator is much more cost-effective. Hence, it is advantageous to leverage simulation data through a transfer learning framework.

In this study, we focus on the task of identifying the target object in a given scenario using natural language instructions for object manipulation.
For instance, given the instruction ``Bring me the book closest to the lamp,'' and a scene in which several books are near the lamp, the robot is expected to specify the book closest to the lamp as the target object.
It is not easy to understand the meaning of human instructions correctly because such instructions are often ambiguous. In the above example, the robot should identify the book closest to the lamp among all the observed books by correctly comprehending the referring expression in the given instruction. Magassouba et al. \cite{magassouba2020multimodal} report cases in which a robot fails to comprehend instructions containing referring expressions.

In this study, we aim to transfer experience gained from \textit{simulation} data to \textit{real-world} data to improve performance in our target task.
In the task of multimodal language understanding for object manipulation, most conventional models have been trained only on real-world data \cite{hatori2018interactively,magassouba2019understanding,ishikawa2021target}. However, such approaches have difficulty in increasing dataset size because building real-world datasets is labor-intensive.
By contrast, collecting training data using a simulator is a significantly more cost-effective approach.
Consequently, we expect that leveraging simulation data within a transfer learning framework will effectively enhance model performance.

\begin{figure}
    \centering
    \includegraphics[width=\linewidth]{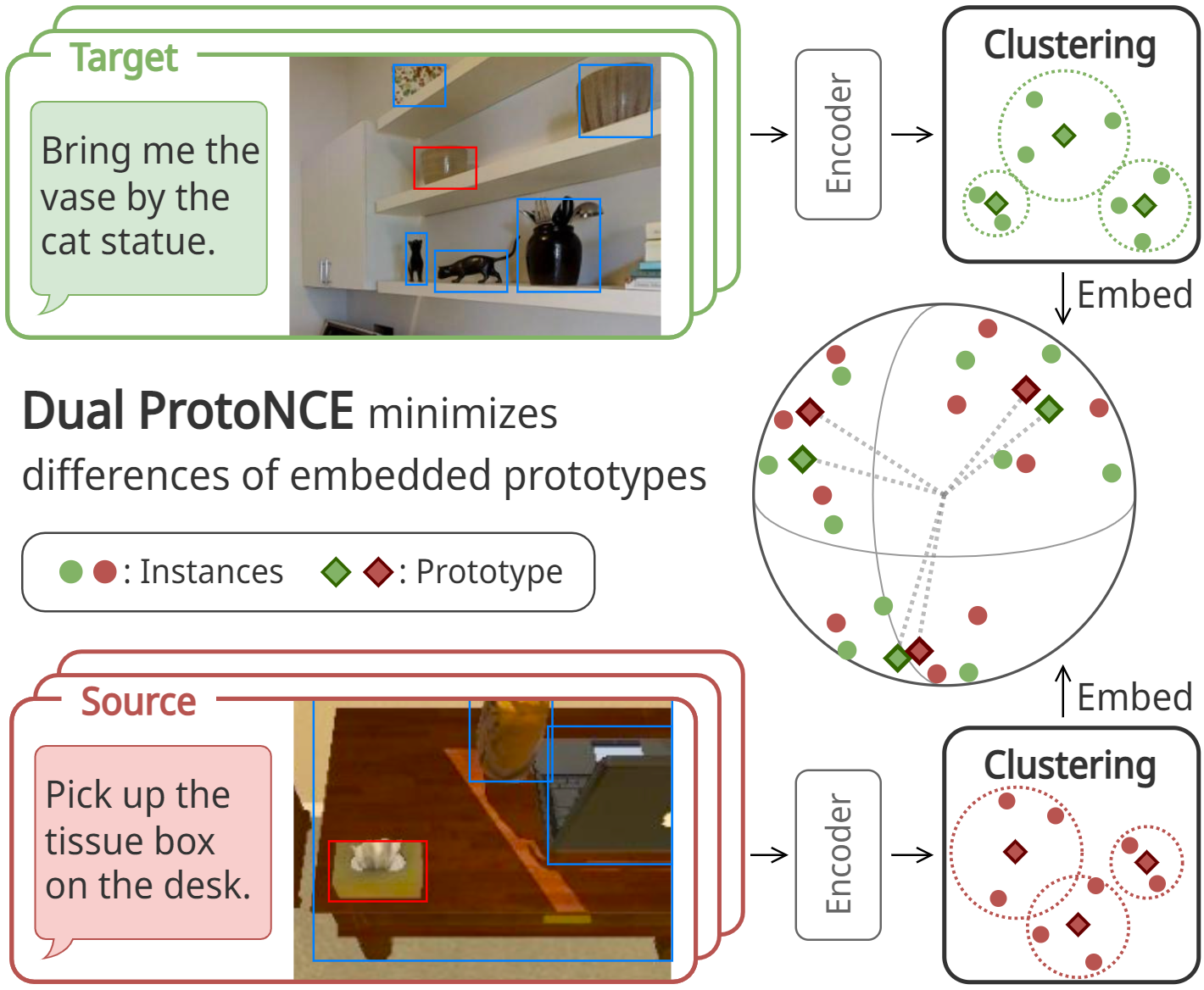}
    \caption{Summary of our method PCTL. Given data from the source and target domains, PCTL aims to alleviate the influence of the domain gap by minimizing Dual ProtoNCE.}
    \label{fig:eye-catch}
\end{figure}

In this paper, we propose Prototypical Contrastive Transfer Learning (PCTL), which is a novel transfer learning approach for the multimodal language understanding task.
PCTL performs contrastive learning between the source and target domain data using our new contrastive loss called Dual ProtoNCE. We expect that PCTL will alleviate the influence of the gap between the two domains by minimizing Dual ProtoNCE.
A summary of our method is shown in Fig. \ref{fig:eye-catch}.

To design Dual ProtoNCE, we extended ProtoNCE \cite{li2021prototypical} to transfer learning. 
ProtoNCE does not simultaneously handle source and target domains because it is not designed for transfer learning. Unlike ProtoNCE, Dual ProtoNCE is designed as a contrastive loss between the source and target domains.
By defining a new contrastive loss between the source and target domains, we expect that Dual ProtoNCE will enable contrastive representation learning to bridge the gap between the two domains.

Our key contributions are as follows:
\begin{itemize}
    \item We introduce transfer learning to the task of identifying target objects in domestic environments according to free-form natural language instructions.
    \item We propose PCTL, which is a novel transfer learning approach for the multimodal language understanding task.
    \item Within PCTL, we develop Dual ProtoNCE, which is a novel contrastive loss generalized for transfer learning.
\end{itemize}

% \vspace{-1.4mm}
\section{Related Work
\label{related}
}
% \vspace{-0.8mm}
\subsection{Multimodal Language Understanding}
Many surveys have been conducted on multimodal language understanding and vision-language pretraining (VLP) \cite{mogadala2021trends,UPPAL2022149,du2022survey,long2022vision,chen2023vlp}.
Uppal et al. \cite{UPPAL2022149} present an overview of the latest trends in research on multimodal language understanding. They consider task formulations, evaluation metrics, model architecture, and other topics, for example, bias and fairness, and adversarial attacks.
Long et al. \cite{long2022vision} describe the general task definition and architecture of recent VLP models. They also discuss vision and language data encoding methods, and the mainstream model structure. They further summarize several essential pretraining and fine-tuning strategies.

Several studies have tackled referring expression comprehension (REC), which is one of the multimodal language understanding tasks \cite{yu2018mattnet,lu2019vilbert,chen2020uniter,gan2020large,kamath2021mdetr,wang2022ofa}.
In the REC task, models should ground a target object in an image described by a referring expression.
Our task formulation is slightly more flexible than that of REC. In detail, we can address cases in which more than one or no target object exists in a given scene by formulating the task as the binary classification of whether or not the candidate object is the target.
Some studies have attempted to build models for specifying the target object using natural language instructions and visual information \cite{magassouba2019understanding,ishikawa2021target}. Specifically, Target-Dependent UNITER\cite{ishikawa2021target} (TDU) uses UNITER-based transformer architecture to model the relationship between text and visual features. Thus, TDU is pretrainable on general-purpose datasets. Additionally, Ishikawa et al. \cite{ishikawa2021target} formulates the task of identifying the target object in a given scenario using natural language instructions for object manipulation as the Multimodal Language Understanding for Fetching Instruction (MLU-FI) task. Ishikawa et al. \cite{ishikawa2022moment} proposes Moment-based Adversarial Training (MAT), which is an adversarial training approach for vision-and-language navigation (VLN) tasks.

\subsection{Datasets}
Several datasets exist for the MLU-FI task. PFN-PIC \cite{hatori2018interactively} is a dataset that consists of images and instructions about objects in the scene. It contains images of approximately 20 commodities in four boxes taken in the real world, with the limitation of a fixed viewpoint. WRS-unialt \cite{ishikawa2021target} is a dataset that consists of images and instructions collected using a simulator. Additionally, these images were observed from various viewpoints.

In this study, we target the MLU-FI task in various real-world indoor environments with scenes observed from various viewpoints. However, to the best of our knowledge, no standard real-world dataset exists for this task. Therefore, we built new datasets by collecting data necessary for the task from datasets used in the VLN task.
Several standard datasets exist for VLN \cite{anderson2018vision,qi2020reverie,ku2020room}.
The Room-to-Room (R2R) dataset \cite{anderson2018vision} is a benchmark dataset for VLN in building-scale 3D environments in the real world. In the R2R navigation task, autonomous agents are required to follow navigation instructions in previously unseen indoor environments.
The Remote Embodied Visual Referring Expression in Real Indoor Environments (REVERIE) dataset \cite{qi2020reverie} is a standard dataset for the VLN task in real indoor environments. The REVERIE task consists of the subtask of navigating to a location where the target object exists, followed by another subtask of identifying the target object.
These VLN datasets were built on the data provided by MatterPort3D \cite{chang2017matterport3d}. MatterPort3D is a large-scale RGB-D dataset for scene understanding in various indoor environments in the real world. The dataset contains 10,800 panoramic views from 194,400 RGB-D images of 90 building-scale scenes with varied annotation, such as segmentation information.

\subsection{Contrastive Learning}
Contrastive learning is an approach used to learn a good data representation in a self-supervised manner. This category of learning strategies aims to align all instances in the embedding space where they are well-separated and locally smooth by leveraging the contrastive loss. 
Various  contrastive learning frameworks have been proposed in the context of representation learning for vision \cite{chen2020simple,he2020momentum,misra2020self}, language \cite{gao2021simcse}, and multimodal models\cite{radford2021learning,jia2021scaling,wu2022data}.

ProtoNCE \cite{li2021prototypical} is a contrastive loss designed to implicitly encode the semantic structure of data into the embedding space by leveraging data prototypes obtained by clustering on embeddings as positive and negative features.
In this study, we develop a novel contrastive loss called Dual ProtoNCE by extending ProtoNCE to transfer learning. Unlike ProtoNCE, we enable contrastive learning across different domains by leveraging data prototypes obtained from clustering for the embedded features of each domain.

% \vspace{-1.0mm}
\section{Problem Statement
\label{sec:problem}
}
% \vspace{-0.8mm}

\subsection{Preliminaries}

The terminology used in this paper is defined as follows:
\begin{itemize}
    \item {\textbf{Target object:}} object referred to in the natural language instruction.
    \item {\textbf{Candidate object:}} object that the model predicts whether it matches the target object or not.
    \item {\textbf{Context objects:}} objects detected by an object detector.
\end{itemize}
We refer to the bounding boxes of the target, candidate, and context objects as the target, candidate, and context regions, respectively.

% \vspace{-0.5mm}
\subsection{Task Formulation}
% In this paper, 
We focus on the MLU-FI task.
In this task, given a natural language instruction, candidate region, and context regions, the model is required to perform the binary classification of whether the candidate object matches the target object or not.

The MLU-FI task is characterized as follows:
\begin{itemize}
    \item {Input:} An instruction, candidate region, and context regions.
    \item {Output:} Predicted probability $p(\hat{y}=1)$. $y$ and $\hat{y}$ denote a label and predicted label, respectively. The condition $y=1$ indicates that the candidate object matches the target object.
\end{itemize}
% We employ classification accuracy as the evaluation metric.

\begin{figure}
    \centering
    \includegraphics[width=\linewidth]{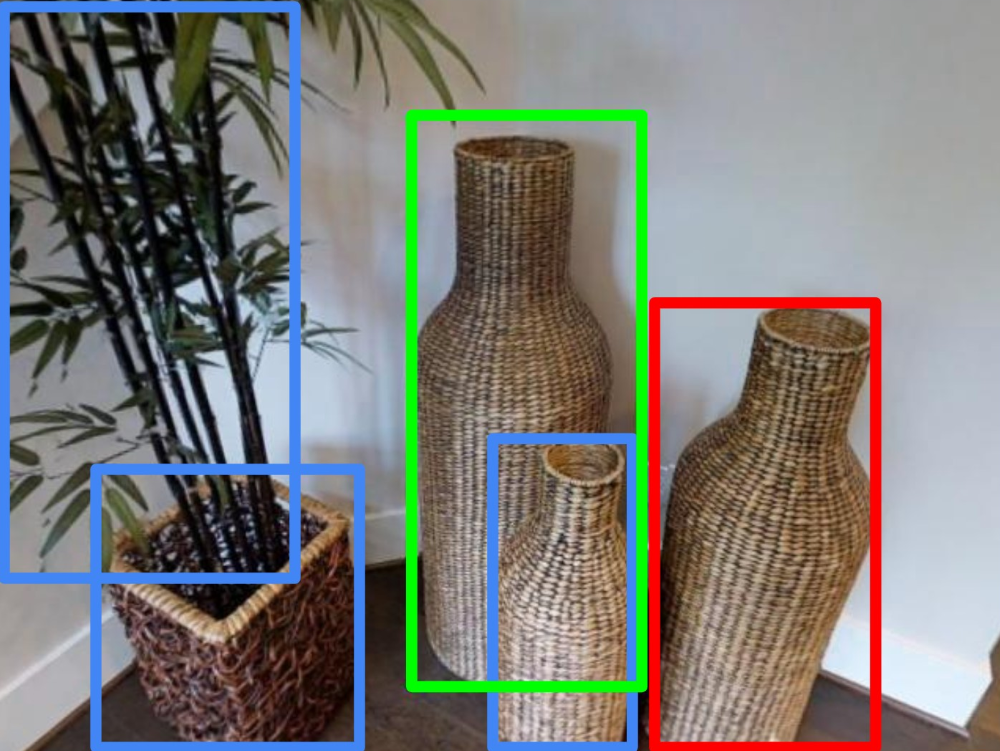}
    \caption{Typical sample of the target task. The instruction is ``Look in the left wicker vase next to the potted plant on the second floor at the foot of the stairs.'' The red and green bounding boxes represent the candidate and target regions, respectively. Note that blue bounding boxes represent the context regions.}
    \label{fig:main_example}
\end{figure}
Fig. \ref{fig:main_example} shows a typical sample of the task. In the sample, the target object is the wicker vase enclosed by the green bounding box. In this case, $p(\hat{y}=1)=0$ because the given candidate object does not match the target object.

We assume that an object detector is used to extract the candidate region and context regions from the image.
It is worth noting that the task is not a multi-class classification task to select a single object from all the objects in the image. The binary classification setting allows us to consider the case in which multiple or no target objects exist in the given image.

For transfer learning, the target samples are collected in the real world, whereas the source samples are collected using a simulator. Every sample consists of a set of an instruction, a candidate region, and context regions. They are collected in indoor environments for the MLU-FI task.

% \vspace{-1mm}
\section{Proposed Method
\label{method}
}
% \vspace{-0.8mm}
% \subsection{Novelty}
In this study, we propose PCTL, which is a novel transfer learning approach for the multimodal language understanding task. Specifically, we introduce Dual ProtoNCE, which is a contrastive loss generalized for transfer learning.
Although we apply our transfer learning approach to the MLU-FI task, our proposed method can be used for transfer learning on other multimodal language understanding tasks.

\begin{figure*}
    \centering
    \includegraphics[width=\linewidth]{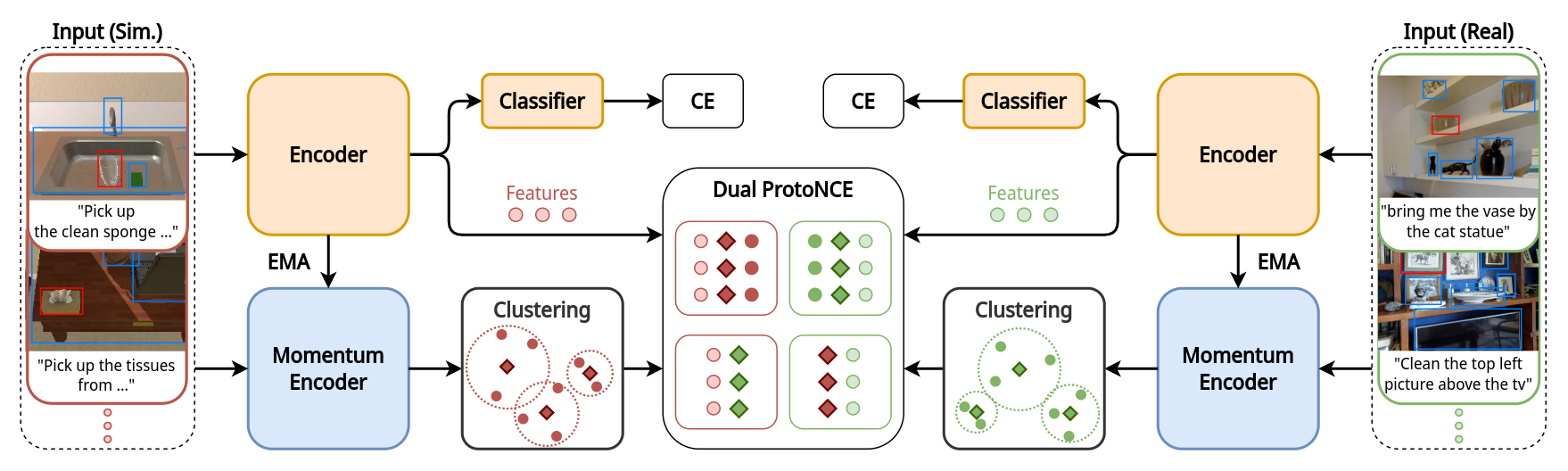}
    \caption{Training framework for PCTL. In our framework, we train two networks: Classifier and Encoder with cross-entropy loss (CE) and Dual ProtoNCE. Dual ProtoNCE uses momentum features and \textit{prototypes} estimated via clustering on them.}
    \label{fig:framework}
\end{figure*}

Fig. \ref{fig:framework} shows an overview of our training framework.
Our overall framework, referred to as PCTL, has three main modules: Encoder, Momentum Encoder, and Clustering Module.

% Our key contributions are the following:
% \begin{itemize}
%     \item We introduce transfer learning to the MLU-FI task.
%     \item We propose PCTL, which is a novel transfer learning approach for the multimodal language understanding task.
%     \item Within PCTL, we develop Dual ProtoNCE -- a novel contrastive loss generalized for transfer learning.
% \end{itemize}

% \subsection{Model Architecture}
% \subsubsection{Input}
\subsection{Input}
We define the input $\bm{x}$ to our model as follows:
\begin{align}
    \bm{x} &= \{\bm{x}_\mathrm{inst}, \bm{x}_\mathrm{cand}, \bm{X}_\mathrm{cont} \}\\
    \bm{X}_\mathrm{cont} &= \{ \bm{x}_\mathrm{cont}^{(i)} \vert i = 1, \dots, N_\mathrm{det} \},
\end{align}
where $\bm{x}_{\mathrm{inst}}$, $\bm{x}_{\mathrm{cand}}$, and $\bm{x}_\mathrm{cont}^{(i)}$ denote a natural language instruction, candidate region, and $i$-th context region, respectively. We use Faster R-CNN to detect $N_\mathrm{det}$ context regions.

It should be noted that we use positional encoding for $\bm{x}_\mathrm{cand}$, $\bm{X}_\mathrm{cont}$, and the text features extracted from $\bm{x}_\mathrm{inst}$. We use a seven-dimensional vector $[x_1, y_1, x_2, y_2, x_2-x_1, y_2-y_1, (x_2-x_1) \cdot (y_2-y_1)]^T$ as positional encoding for $\bm{x}_\mathrm{cand}$ and $\bm{X}_\mathrm{cont}$, where $(x_1, y_1)$ and $(x_2, y_2)$ are the coordinates of the top-left and bottom-right corners, respectively. Additionally, we assume that $x_1$, $x_2$, $y_1$, and $y_2$ are normalized by the width and height of the input image.

% We tokenize $\bm{x}_\mathrm{inst}$ using WordPiece\cite{wu2016google} and convert them to text features by Text Embedder of Target-Dependent UNITER\cite{ishikawa2021target}. With regard to visual features, we extract $\bm{x}_\mathrm{cand}$ and $\bm{x}_\mathrm{cont}^{(i)}$ from the output of the fc7 layer following the RoI pooling layer of Faster R-CNN\cite{ren2017faster} which has a ResNet50 backbone network.

% \subsubsection{Encoder}
\subsection{Encoder}
Encoder $f_{\bm \theta}$ is parameterized by $\bm{\theta}$. The structure of $f_{\bm \theta}$ follows TDU\cite{ishikawa2021target} and has three main parts: Text Embedder, Image Embedder, and Multi-Layer Transformer. The Text Embedder tokenizes $\bm{x}_\mathrm{inst}$ using WordPiece\cite{wu2016google} and converts them to text features. The Image Embedder embeds $\bm{x}_\mathrm{cand}$ and $\bm{x}_\mathrm{cont}^{(i)}$ into visual features. The Multi-Layer Transformer takes text and visual features as input and models the relationship between them. The output is the final hidden vector of the Multi-Layer Transformer corresponding to the input feature, $\bm{x}_\mathrm{cand}$.

Hereafter, we refer to a sample of the source domain as $\left(\bm{x}_\mathrm{s}, y_\mathrm{s}\right)$. Similarly, $\left(\bm{x}_\mathrm{t}, y_\mathrm{t}\right)$ denotes that of the target domain. The input and output of $f_{\bm \theta}$ are denoted as follows:
\begin{align}
	\bm{u} &= f_{\bm \theta}(\bm{x}_\mathrm{s}) \in \mathbb{R}^{768}\\
	\bm{v} &= f_{\bm \theta} (\bm{x}_\mathrm{t}) \in \mathbb{R}^{768},
\end{align}
where $\bm{u}$ and $\bm{v}$ denote the feature vector of the source sample $\bm{x}_\mathrm{s}$ and that of the target sample $\bm{x}_\mathrm{t}$. They are used for $k$-means clustering and the loss function. Classifier $g$ consists of a two-layer MLP and softmax function. $g$ takes the output of $f_{\bm \theta}$ and calculates the predicted probability,
\begin{align}
    p(\hat{y}=1) = g(f_{\bm \theta}(\bm x)).
\end{align}

% \subsubsection{Momentum Encoder}
\subsection{Momentum Encoder}
The Momentum Encoder $f_{\bm \theta'}$ has the same structure as $f_{\bm \theta}$. $f_{\bm \theta'}$ is parametrized by $\bm \theta'$, which is modeled as a moving average of $\bm \theta$. Specifically, we update $\bm \theta'$ as follows:
\begin{align}
    {\bm \theta'} \leftarrow \gamma {\bm \theta'} + (1 - \gamma) {\bm \theta},
\end{align}
where $\gamma$ denotes a smoothing coefficient. Similar to $f_{\bm \theta}$, we denote the output of $f_{\bm \theta'}$ as follows:
\begin{align}
	\bm{u}' &= f_{\bm \theta'} (\bm{x}_\mathrm{s}) \in \mathbb{R}^{768}\\
	\bm{v}' &= f_{\bm \theta'} (\bm{x}_\mathrm{t}) \in \mathbb{R}^{768}.
\end{align}

\subsection{Clustering}
The Clustering Module performs $k$-means clustering on $\bm{u}'$ and $\bm{v}'$ $M$ times at the beginning of each epoch.
We define the $i$-th \textit{prototype} as the centroid of the $i$-th cluster. Suppose $\bm{c}_i^{(m)}$ and $\bm{d}_i^{(m)}$ denote the \textit{prototypes} of the $i$-th cluster that results from the $i$-th clustering step on $\bm{u}'$ and $\bm{v}'$, respectively. $k^{(m)}$ denotes the number of clusters in the $m$-th clustering step.

% \subsection{Contrastive Transfer Learning}
\subsection{Contrastive Transfer Learning}
The Contrastive Transfer Learning aims to train the model to bridge the gap between the source and target domains by minimizing Dual ProtoNCE, which is a novel contrastive loss generalized for transfer learning.

\subsubsection{InfoNCE}
Given a training set $\bm{A}=\left\{\bm{a}_1, \bm{a}_2, \dots, \bm{a}_n \right\}$, unsupervised representation learning is designed to train the encoder $f$ that maps $\bm A$ to embeddings $\bm{Z} = \left\{ \bm{z}_1, \bm{z}_2, \dots, \bm{z}_n \right\}$ so that $\bm{z}_i = f(\bm{a}_i)$ best describes $\bm{a}_i$. Contrastive learning achieves this goal by minimizing the contrastive loss typified by InfoNCE\cite{oord2018representation, he2020momentum}. Let $\bm{z}_i$, $\bm{z}'_i$, and $\left\{ \bm{z}'_j \mid j = n+1, n+2, \dots, n+r \right\}$ be anchor, positive, and $r$ negative embeddings, respectively. Then, InfoNCE is defined as
\begin{align}
    \mathcal{L}_\mathrm{InfoNCE} = \sum_{i=1}^n - \log\frac{\exp(\bm{z}_i \cdot \bm{z}'_i / \tau)}{\sum_{j \in J} \exp(\bm{z}_i \cdot \bm{z}'_j / \tau)},
\end{align}
where $J = \left\{ i, n+1, n+2, \dots, n+r \right\}$. We set $\tau$ as a learnable temperature parameter following \cite{radford2021learning}.
% We initialize $1 / \tau$ by $0.07$ and clip $1 / \tau$ to prevent scaling the logits by more than 100 which we found necessary to prevent training instability.
We initialize $1/\tau$ to $0.07$, and apply clipping to avoid scaling the logits beyond $100$. 

\subsubsection{ProtoNCE}
ProtoNCE\cite{li2021prototypical} is a contrastive loss designed to push embeddings and their assigned prototypes together while pushing those and other prototypes apart. Let $\bm{h}_i^{(m)}$ be the $i$-th \textit{prototype} obtained in $m$-th clustering w.r.t. $\bm{z}'$. ProtoNCE is defined by the following equation:
\begin{align}
    \begin{split}
        \mathcal{L}_\mathrm{ProtoNCE} =& \sum_{i=1}^n - \left(\log\frac{\exp(\bm{z}_i \cdot \bm{z}'_i / \tau)}{\sum_{j \in J} \exp(\bm{z}_i \cdot \bm{z}'_j / \tau)}\right. \\
        &\hspace{-30pt} + \left.\frac{1}{M} \sum_{m=1}^M \log \frac{\exp(\bm{z}_i \cdot \bm{h}_s^{(m)} / \phi_s^{(m)} )}{ \sum_{j \in J'} \exp(\bm{z}_i \cdot \bm{h}_j^{(m)} / \phi_j^{(m)})}\right), \\
        J' \subset \{ &1, 2, \dots, k^{(m)} \}, s \in J', r' = \vert J'\setminus \{s\} \vert.
    \end{split}
\end{align}
In the above expression, $\bm{h}_s$ denotes the positive \textit{prototype} closest to $\bm{z}_i$, whereas $\left\{ \bm{h}_j^{(m)} \mid j \in J' \setminus \{s\} \right\}$ denotes $r'$ negative \textit{prototypes} randomly selected from all the \textit{prototypes} of the $m$-th clustering result except the positive prototype, $\bm{h}_s$. $\phi$ indicates the level of instance-to-\textit{prototype} concentration for each cluster. Hereafter, we call $\phi$ the concentration factor. Concentration factor $\phi_i$ for the \textit{prototype} $\bm{h}_i$ of the $i$-th cluster $\bm{C}_i$ is defined as
\begin{align}
    \phi_i = \frac{\sum_{\bm{z}' \in \bm{C}_i} \Vert \bm{z}' - \bm{h}_i \Vert_2}{\vert \bm{C}_i \vert \log (\vert \bm{C}_i \vert + \alpha)} \cdot \frac{\tau'}{\sum_{j=1}^{k^{(m)}} {\phi_j / k^{(m)}}},
\end{align}
where $\vert \bm{C}_i \vert$ is the number of instances $\bm{z}_i$ assigned to $\bm{C}_i$, and $\alpha$ denotes a smoothing parameter used to avoid the divergence of $\phi_i$ for the small cluster. We normalize $\phi$ over all the clusters so that they have a mean of $\tau'$.

\subsubsection{Dual ProtoNCE}
The Dual ProtoNCE loss is a novel contrastive loss expanded from ProtoNCE.
% $\mathcal{L}_\mathrm{DualProtoNCE}$
It is defined as the summation of the two losses, Intra-Domain Loss $\mathcal{L}_\mathrm{Intra}$ and Inter-Domain Loss $\mathcal{L}_\mathrm{Inter}$, as follows:
\begin{align}
    \mathcal{L}_\mathrm{DualProtoNCE} = \mathcal{L}_\mathrm{Intra} + \mathcal{L}_\mathrm{Inter}.
\end{align}

We first compute $\mathcal{L}_\mathrm{Intra}$ by applying ProtoNCE to source samples and target samples independently:
\begin{align}
    \begin{split}
        \mathcal{L}_\mathrm{Intra} &= \mathcal{L}_\mathrm{Target} + \mathcal{L}_\mathrm{Source}
    \end{split}\\
    \begin{split}
        \mathcal{L}_\mathrm{Target} &= \sum_{i=1}^n -\left( \log \frac{\exp(^{\dagger}\bm{v}_i \cdot {^{\dagger}\bm{v}'_i} / \tau)}{\sum_{j \in J} \exp(^{\dagger}\bm{v}_i \cdot {^{\dagger}\bm{v}'_j} / \tau)} \right.\\
        &\hspace{-20pt} + \left. \frac{1}{M}\sum_{m=1}^M \log\frac{\exp(^{\dagger}\bm{v}_i \cdot {^{\dagger}\bm{c}_s^{(m)}}/\phi_s^{(m)})}{\sum_{j \in J'} \exp(^{\dagger}\bm{v}_i \cdot {^{\dagger}\bm{c}_j^{(m)}} / \phi_j^{(m)})} \right)
    \end{split}\\
    \begin{split}
        \mathcal{L}_\mathrm{Source} &= \sum_{i=1}^n -\left( \log \frac{\exp(^{\dagger}\bm{u}_i \cdot {^{\dagger}\bm{u}'_i} / \tau)}{\sum_{j \in J} \exp(^{\dagger}\bm{u}_i \cdot {^{\dagger}\bm{u}'_j} / \tau)} \right. \\
        &\hspace{-20pt} + \left. \frac{1}{M}\sum_{m=1}^M \log\frac{\exp(^{\dagger}\bm{u}_i \cdot {^{\dagger}\bm{d}_s^{(m)}}/\varphi_s^{(m)})}{\sum_{j \in J'}\exp(^{\dagger}\bm{u}_i \cdot {^{\dagger}\bm{d}_j^{(m)}} / \varphi_j^{(m)})} \right),
    \end{split}
\end{align}
where $^{\dagger}\bm{a}$ represents $\bm{a} \left/ {\Vert \bm{a} \Vert_2} \right.$ and $\phi$ and $\varphi$ denote the concentration factors of $\bm{c}$ and $\bm{d}$, respectively.

Next, we compute $\mathcal{L}_\mathrm{Inter}$ to bridge the gap between the two domains. $\mathcal{L}_\mathrm{Inter}$ is defined as follows:
\begin{align}
	\mathcal{L}_\mathrm{Inter} = \mathcal{L}_\mathrm{S2T} + \mathcal{L}_\mathrm{T2S},
\end{align}
where $\mathcal{L}_\mathrm{S2T}$ is the contrastive loss defined between source domain features $\bm{u}$ and the prototypes of target domain $\bm{c}$, and $\mathcal{L}_\mathrm{T2S}$ is similarly defined between $\bm v$ and $\bm d$. They are expressed as
\begin{align}
    \begin{split}
        \mathcal{L}_\mathrm{S2T} =& -\frac{1}{M}\sum_{i=1}^n\sum_{m=1}^M\\
        &\left( \log \frac{\exp(^\dagger \bm{u}_i \cdot {^\dagger \bm{c}_s^{(m)}} / \phi_s^{(m)})}{\sum_{j \in J'} \exp(^\dagger \bm{u}_i \cdot {^\dagger \bm{c}_j^{(m)}} / \phi_j^{(m)})} \right),
    \end{split}\\
    \begin{split}
        \mathcal{L}_\mathrm{T2S} =& -\frac{1}{M}\sum_{i=1}^n\sum_{m=1}^M\\
        &\left( \log \frac{\exp(^\dagger \bm{v}_i \cdot {^\dagger \bm{d}_s^{(m)}} / \varphi_s^{(m)})}{\sum_{j \in J'} \exp(^\dagger \bm{v}_i \cdot {^\dagger \bm{d}_j^{(m)}} / \varphi_j^{(m)})} \right).
    \end{split}
\end{align}

Let $\mathcal{L}_\mathrm{CE}$ and $\lambda$ be the cross-entropy loss and a hyperparameter, respectively. Our overall loss function $\mathcal{L}$ is defined as,
\begin{align}
    \begin{split}
        \mathcal{L} &= \lambda \mathcal{L}_\mathrm{DualProtoNCE} + \mathcal{L}_\mathrm{t} + \mathcal{L}_\mathrm{s}
    \end{split}\\
    \begin{split}
        \mathcal{L}_\mathrm{t} &= \left.\sum_{i=1}^n\right(\mathcal{L}_\mathrm{CE}(g(f_\theta(\bm{x}_\mathrm{t}^{(i)})), y_\mathrm{t}^{(i)})\\
        &\hspace{40pt} +\left.\frac{1}{M}\sum_{m=1}^M \mathcal{L}_\mathrm{CE}(g(\bm{c}_s^{(m)}), y_\mathrm{t}^{(i)}) \right)
    \end{split}\\
    \begin{split}
        \mathcal{L}_\mathrm{s} &= \left.\sum_{i=1}^n \right(\mathcal{L}_\mathrm{CE}(g(f_\theta(\bm{x}_\mathrm{s}^{(i)})), y_\mathrm{s}^{(i)})\\
        &\hspace{40pt} +\left.\frac{1}{M}\sum_{m=1}^M \mathcal{L}_\mathrm{CE}(g(\bm{d}_s^{(m)}), y_\mathrm{s}^{(i)}) \right),
    \end{split}
\end{align}
where $\bm{c}_s^{(m)}$ and $\bm{d}_s^{(m)}$ are the \textit{prototypes} closest to the embedding $f_\theta(\bm{x}_\mathrm{t}^{(i)})$ and $f_\theta(\bm{x}_\mathrm{s}^{(i)})$, respectively.

\begin{table}[!b]
    \normalsize
    \renewcommand*{\arraystretch}{1.25}
    \newcommand*{\bhline}[1]{\noalign{\hrule height #1}}
    \caption{Experimental setup.}
    \centering
    \begin{tabular}{lll}
    \bhline{0.6pt}
    \multirow{2}{*}{PCTL}\hspace{10pt} & \multirow{2}{*}{$\mathcal{L}_\mathrm{DualProtoNCE}$} & $(r, r')=(32, 32),$ \\
     & & $\: k^{(1)} = 64,\: \lambda = 1/32$\\ \cline{2-3}
                          & Concentration, $\phi$ & $\tau' = 0.2,\: \alpha = 10$ \\ \hline
    \multicolumn{2}{l}{Transformer}     & \#L: 12, \#H: 768, \#A: 12 \\ \hline
    \multicolumn{2}{l}{Optimizer}     & SGD w/ momentum 0.9 \\ \hline
    \multicolumn{2}{l}{Learning rate (LR)}     & $8 \times 10^{-4}$  \\ \hline
    \multicolumn{2}{l}{
    \multirow{2}{12em}{LR \hspace{9em} for Multi-Layer Transformer}
    }
     & \multirow{2}{*}{$8 \times 10^{-5}$} \\
     & & \\\hline
    \multicolumn{2}{l}{Batch size}      & 64  \\ \hline
    \multicolumn{2}{l}{\#Epoch}      & 30  \\
    \bhline{0.6pt}
    \end{tabular}
    \label{tab:exp_setup}
\end{table}

\section{Experiments
\label{exp}
}

\subsection{Datasets}
To validate our model in real-world domestic environments, we built a new dataset called REVERIE-fetch.
This is because no standard real-world dataset exists for the MLU-FI task, to the best of our knowledge.
To construct such a dataset, we collected images and natural language instructions based on the REVERIE dataset \cite{qi2020reverie}, which is a standard dataset for VLN in real-world indoor environments. Note that this dataset is not directly applicable to our task.

We first collected the cubemaps \cite{cubemap} of the goal points provided in the original dataset because the target object is placed at the goal point of the navigation task in the REVERIE task. Then, we extracted images in which target objects existed from the collected cubemaps.
Over 1,000 annotators collected the instructions in the REVERIE dataset using Amazon Mechanical Turk. The annotators viewed an animation of the route and a randomly highlighted target object via the interactive 3D WebGL simulator. Then, they were asked to provide instructions to find and manipulate the target object.

Regarding the source-domain datasets, we extracted the source samples from the ALFRED dataset \cite{shridhar2020alfred} and built the ALFRED-fetch-b dataset. The ALFRED dataset is a standard dataset for VLN with object manipulation.
It includes 25,743 English instructions that describe 8,055 expert demonstrations.
It contains multiple sequential subgoals that constitute the given goal, an instruction for each subgoal, and images observed from the agent’s first-person views at each timestep of ground-truth behavior.
The new ALFRED-fetch-b dataset consists of instructions and images from the training set of the original ALFRED dataset. They were collected in a scenario in which the subgoal was to pick up an object. In this study, we gathered the images from scenes just before the picking action.

As a preprocessing step for data extracted from the REVERIE and ALFRED datasets, we extracted the candidate and context regions from the images using Faster R-CNN \cite{ren2017faster} to create samples. We labeled those with a GIoU \cite{rezatofighi2019generalized} greater than 0.80 of their target and candidate regions as positive samples and those with a GIoU less than 0.45 as negative samples. The target regions were provided in the original datasets. In the test set of REVERIE-fetch, we manually removed inappropriate samples caused by misdetection.

The REVERIE-fetch dataset consists of 10,243 image-instruction pairs with a vocabulary size of 1958 words, a total of 188,965 words, and an average sentence length of 18.4 words. The ALFRED-fetch-b dataset similarly consists of 34,286 samples of image-instruction pairs and a total of 399,964 words. Its vocabulary size and average sentence length are 1,558 and 11.7 words, respectively.
The REVERIE-fetch dataset includes 8,302, 994, and 947 samples in the training, validation, and test sets, respectively. We built the training set with data collected from the training set and the seen split of the validation set of the REVERIE dataset. The validation and test sets consist of data collected from the unseen split of the validation set of the REVERIE dataset.
The ALFRED-fetch-b dataset includes 27,492; 3,470; and 3,324 samples in the training, validation, and test sets, respectively. We collected these samples from the training set of the ALFRED dataset.
It should be mentioned that there were no overlaps among the training, validation, and test sets for either dataset.
We used the training set to train our model and the validation set to tune the hyperparameters. We evaluated our model on the test set of the REVERIE-fetch dataset.

\begin{table}[!t]
    \normalsize
    \renewcommand*{\arraystretch}{1.25}
    \newcommand*{\bhline}[1]{\noalign{\hrule height #1}}
    \centering
    \captionsetup{justification=centering}
    \caption{\\Quantitative results on the REVERIE-fetch dataset.}
    \begin{tabular}{ m{0.40\linewidth} Wc{0.3\linewidth} }
    \bhline{0.6pt}
        \hspace{15pt} Method & Accuracy [\%]  \\ \hline
        \hspace{15pt} Target domain only\hspace{15pt} & $73.0\pm1.87$ \\
        \hspace{15pt} Fine-tuning & $73.4\pm11.8$ \\
        \hspace{15pt} MCDDA+\cite{saito2018maximum} & $74.9\pm3.94$ \\
        \hspace{15pt} Ours & $\mathbf{78.1\pm2.49}$ \\
    \bhline{0.6pt}
    \end{tabular}
    \label{tab:quantitative1}
\end{table}

\subsection{Experimental Setup}
Table \ref{tab:exp_setup} summarizes the experimental setup. Note that \#L, \#H, and \#A denote the number of layers, hidden size, and number of attention heads in the Multi-Layer Transformer, respectively.

Our model had roughly 110 million trainable parameters.
We trained our model on a GeForce RTX 3090 with 24GB of memory and an Intel Core i9-10900KF with 64GB of memory.
It took 2 hours to train our model. The inference time was approximately 59.3 milliseconds for one sample.
We evaluated the value of the loss $\mathcal{L}_\mathrm{CE}$ of the model for every epoch on the validation set of REVERIE-fetch. We used the test set accuracy of the REVERIE-fetch dataset when the value of $\mathcal{L}_\mathrm{CE}$ on the validation set of REVERIE-fetch was minimized.

\subsection{Quantitative Results}
We conducted experiments to compare the proposed and baseline methods.
Table \ref{tab:quantitative1} shows the accuracy on the REVERIE-fetch dataset. The right column shows the means and standard deviations over five trials.
In this experiment, we used accuracy as the evaluation metric because the numbers of positive and negative samples were almost balanced. 

We set the following three baseline settings:
\begin{enumerate}
    \setlength{\leftskip}{4pt}
    \item[(i)] Target domain only: We trained the model only on target samples.
    \item[(ii)] Fine-tuning: We performed pretraining on the source samples and fine-tuned the model with the target samples.
    \item[(iii)] MCDDA+: We extended the Maximum Classifier Discrepancy for Domain Adaptation \cite{saito2018maximum} (MCDDA) and applied it to a supervised transfer learning setting.
\end{enumerate}
We set up baselines (i) and (ii) to compare our approach with the approach in which no data from the source domain was used and the approach in which pretraining was performed on data from the source domain, respectively. As reported in \cite{saito2018maximum}, MCDDA performs well as an unsupervised transfer learning method on image classification. Therefore, we extended MCDDA to a supervised transfer learning setting and used it as baseline (iii). We called this extended method MCDDA+.

As listed in Table \ref{tab:quantitative1}, our method achieved an accuracy of 78.1\%, whereas the accuracy of baselines (i), (ii), and (iii) were 73.0\%, 73.4\%, and 74.9\%, respectively. Therefore our method outperformed all the baselines (i), (ii), and (iii) by 5.1, 4.7, and 3.2 points in terms of accuracy, respectively.
The performance difference between baseline (i) and our method was statistically significant (p-value was lower than 0.01).

\subsection{Ablation Studies}
We conducted ablation studies to investigate the contribution of $M$ and the combination of $k^{(m)}$ to performance.
Specifically, we set the following conditions:
% \begin{enumerate}
%     \item[(i)] $M=1, k^{(1)} = 32$
%     \item[(ii)] $M=2, \left( k^{(1)}, k^{(2)} \right) = \left( 64, 128 \right)$
%     \item[(iii)] $M=3, \left( k^{(1)}, k^{(2)}, k^{(3)} \right) = \left( 64, 128, 256 \right)$
%     \item[(iv)] $M=4, \left( k^{(1)}, k^{(2)}, k^{(3)}, k^{(4)} \right) = \left( 64, 128, 256, 512 \right)$
% \end{enumerate}

% \begin{table}[!b]
%     \normalsize
%     \renewcommand*{\arraystretch}{1.25}
%     \newcommand*{\bhline}[1]{\noalign{\hrule height #1}}
%     \centering
%     \caption{\\Quantitative results of the ablation studies.}
%     \begin{tabular}{ Wl{0.35\linewidth} Wc{0.35\linewidth} }
%     \bhline{0.6pt}
%         \hspace{20pt} Condition & Accuracy [\%] \hspace{3pt} \\ \hline
%         \hspace{20pt} Ours & $\mathbf{78.1\pm2.49}$ \hspace{3pt} \\
%         \hspace{20pt} (i) & $77.1\pm1.55$ \hspace{3pt} \\
%         \hspace{20pt} (ii) & $75.2\pm1.24$ \hspace{3pt} \\
%         \hspace{20pt} (iii) & $71.7\pm10.3$ \hspace{3pt} \\
%         \hspace{20pt} (iv) & $75.6\pm2.67$ \hspace{3pt} \\
%     \bhline{0.6pt}
%     \end{tabular}
%     \label{tab:ablation}
% \end{table}

% Update
\begin{enumerate}
    \setlength{\leftskip}{8pt}
    \item[(i)-a] $M=1, k^{(1)} = 33$
    \item[(i)-b] $M=1, k^{(1)} = 64$
    \item[(i)-c] $M=1, k^{(1)} = 128$
    \item[(ii)] $M=3, \left( k^{(1)}, k^{(2)}, k^{(3)} \right) = \left( 64, 128, 256 \right)$
\end{enumerate}
% It is worth noting that we chose $k^{(1)} = 33$ for the condition (i)-a, because $33$ is the minimum value for $k^{(1)}$ where we choose the $r'=32$ negative prototypes and a positive prototype from total of $k^{(1)}$ prototypes.
It is worth noting that we chose $k^{(1)} = 33$ for condition (i)-a because it is the minimum value of $k^{(1)}$ that allows us to select 32 negative prototypes ($r'=32$) and one positive prototype from a total of $k^{(1)}$ prototypes.

\begin{table}[!b]
    \normalsize
    \renewcommand*{\arraystretch}{1.25}
    \newcommand*{\bhline}[1]{\noalign{\hrule height #1}}
    \centering
    \caption{\\Quantitative results of the ablation studies.}
    \begin{tabular}{ Wl{0.35\linewidth} Wc{0.35\linewidth} }
    \bhline{0.6pt}
        \hspace{20pt} Condition & Accuracy [\%] \hspace{3pt} \\ \hline
        \hspace{20pt} Ours & $\mathbf{78.1\pm2.49}$ \hspace{3pt} \\
        \hspace{20pt} (i)-a & $75.6\pm1.96$ \hspace{3pt} \\
        \hspace{20pt} (i)-b & $73.7\pm2.92$ \hspace{3pt} \\
        \hspace{20pt} (i)-c & $77.4\pm1.96$ \hspace{3pt} \\
        \hspace{20pt} (ii) & $71.7\pm10.3$ \hspace{3pt} \\
    \bhline{0.6pt}
    \end{tabular}
    \label{tab:ablation}
\end{table}

\begin{figure*}[!t]
    \centering
    \begin{minipage}[t]{0.32\linewidth}
        \centering
        \includegraphics[width=\linewidth]{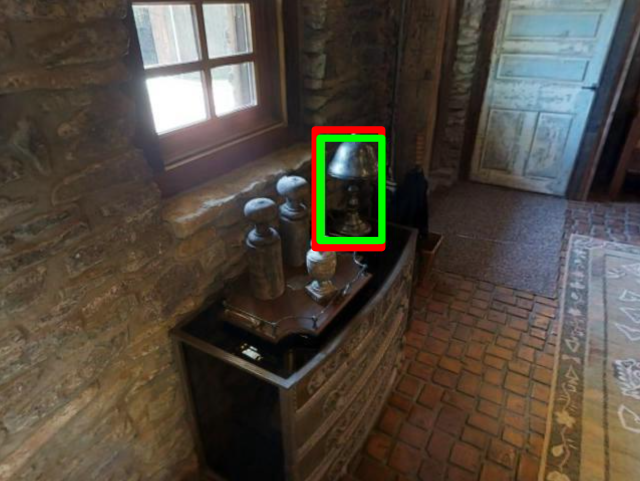}
        \subcaption{\setlength{\leftskip}{1.5mm}\setlength{\rightskip}{1.5mm}``Go down the stairs to the lower balcony area and turn off the lamp on the dresser.''}
    \end{minipage}
    \begin{minipage}[t]{0.32\linewidth}
        \centering
        \includegraphics[width=\linewidth]{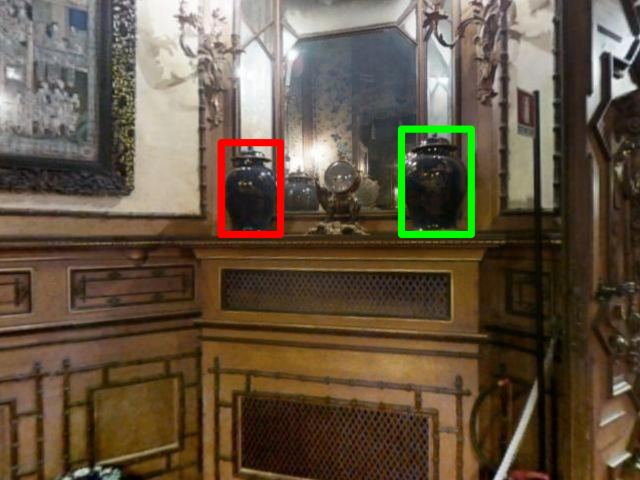}
        \subcaption{\setlength{\leftskip}{1.5mm}\setlength{\rightskip}{1.5mm}``Go to the lounge on the first level where the red carpet is and move the black vase to the right of the mirror.''}
    \end{minipage}
    \begin{minipage}[t]{0.32\linewidth}
        \centering
        \includegraphics[width=\linewidth]{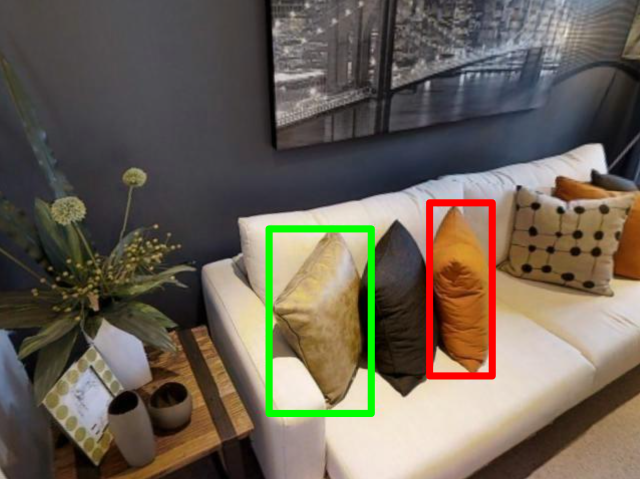}
        \subcaption{\setlength{\leftskip}{1.5mm}\setlength{\rightskip}{1.5mm}``Fluff the light silver pillow on the smaller couch in the living room''}
    \end{minipage}
    \caption{Qualitative results on the REVERIE-fetch dataset. Panels (a), (b), and (c) show the true positive, true negative, and false positive cases, respectively. The red and green bounding boxes indicate the candidate and target regions, respectively. For panels (a) and (b), our method correctly predicted whether the candidate object matched the target object or not by successfully comprehending the instruction and scene.
}
    \label{fig:qualitative1}
\end{figure*}

% Table \ref{tab:ablation} lists the quantitative result of the ablation study. The accuracy column shows the means and standard deviations over five trials. As described in Table \ref{tab:ablation}, PCTL achieved accuracy of 78.1\%, while the accuracy of conditions (i), (ii), (iii), and (iv) were 77.1\%, 75.2\%, 71.7\%, and 75.6\%, respectively.
% This indicates that our method achieved the highest accuracy under all the ablation conditions for $k^{(m)}$ and $M$. Moreover, this result says that decrease or increase of $k^{(m)}$ makes the performance worse.
Table \ref{tab:ablation} lists the quantitative results of the ablation study. The accuracy column shows the means and standard deviations over five trials. As described in Table \ref{tab:ablation}, PCTL achieved accuracy of 78.1\%, whereas the accuracy under conditions (i)-a, (i)-b, (i)-c, and (ii) were 75.6\%, 73.7\%, 77.4\%, and 71.7\%, respectively.
This indicates that our method achieved the highest accuracy under all the ablation conditions for $k^{(m)}$ and $M$. Moreover, this result indicates that a decrease or increase of $k^{(m)}$ reduced performance.

\subsection{Qualitative Results}

Qualitative results are shown in Fig. \ref{fig:qualitative1}.
Fig. \ref{fig:qualitative1} (a) shows a successful sample. The instruction was ``Go down the stairs to the lower balcony area and turn off the lamp on the dresser.'' The target object is the lamp on the dresser. Our method correctly predicted that the candidate object matched the target object, whereas the baseline (i) wrongly predicted that the candidate object did not match the target object.
Fig. \ref{fig:qualitative1} (b) shows another successful sample. The given instruction was ``Go to the lounge on the first level where the red carpet is and move the black vase to the right of the mirror.'' The target object was the vase on the right side of the mirror. Our method successfully identified the candidate object as a different object from the target object, whereas baseline (i) failed to do this.

Fig. \ref{fig:qualitative1} (c) shows a failed sample. The instruction was ``Fluff the light silver pillow on the smaller couch in the living room.'' The target object was the pillow on the left side of the couch. Our method incorrectly predicted that the candidate object matched the target object.

\subsection{Error Analysis and Discussion}

% \begin{table}[!ht]
%     \normalsize
%     \renewcommand*{\arraystretch}{1.50}
%     \caption{Categorization of failed samples}
%     \centering
%     \begin{tabular}{l|l|l|l|l|l|l|l}
%     \hline
%         Error Type & CE & ML & SR & AI & AE & SO & MO \\ \hline
%         \#Error & 43 & 17 & 14 & 11 & 10 & 3 & 2 \\ \hline
%     \end{tabular}
%     \label{tab:error}
% \end{table}
\begin{table}[b]
    \normalsize
    \renewcommand*{\arraystretch}{1.25}
    \newcommand*{\bhline}[1]{\noalign{\hrule height #1}}
    \caption{Categorization of failed samples.}
    \centering
    \begin{tabular}{ m{0.65\linewidth} Wc{0.20\linewidth} }
    \bhline{0.6pt}
        \hspace{15pt} Error Type & \#Error \\ \hline
        \hspace{15pt} Comprehension Error\hspace{40pt} & 43 \\
        \hspace{15pt} Missing Landmark & 17 \\
        \hspace{15pt} Small Region & 14 \\
        \hspace{15pt} Ambiguous Instruction & 11 \\
        \hspace{15pt} Annotation Error & 10 \\
        \hspace{15pt} Severe Occlusion & 3 \\
        \hspace{15pt} Multiple Objects & 2 \\
    \bhline{0.6pt}
    \end{tabular}
    \label{tab:error}
\end{table}

% \begin{table}[!ht]
%     \normalsize
%     \renewcommand*{\arraystretch}{1.50}
%     \caption{confusion matrix}
%     \centering
%     \begin{tabular}{l|l|l|l}
%     \hline
%         TP & FP & FN & TN \\ \hline
%         371 & 65 & 126 & 385 \\ \hline
%     \end{tabular}
%     \label{tab:confusion_matrix}
% \end{table}
% Table XXX shows the confusion matrix on the test set w.r.t our method.
The results consisted of 371, 65, 126, and 385 samples for true positives, false positives (FP), false negatives (FN) and true negatives, respectively. Thus, there were 191 samples for the failed cases. We randomly selected 50 FP and 50 FN samples to analyze the causes of errors. Table \ref{tab:error} categorizes these samples. We classified the causes of errors into the following seven types:
\begin{itemize}
    \item {Comprehension Error (CE):} The model failed to process the visual information and instruction correctly. This class includes cases in which the model failed to comprehend the given referring expression or correctly specify the object to which the textual information in the instruction referred.
    \item {Missing Landmark:} The given image did not contain the visual information w.r.t. the referring expression. For example, the model failed to predict the chair nearest the kitchen because the instruction had the referring expression, ``nearest the kitchen,'' but the given image did not contain a kitchen.  % Name: [仮]
    \item {Small Region:} The model failed to specify the target object because the target region was smaller than 1\% of the entire image area.
    \item {Ambiguous Instruction:} The given instruction was ambiguous; hence, the model failed to specify the target object.
    \item {Annotation Error:} Annotation errors occurred in the bounding boxes and/or instructions.
    \item {Severe Occlusion:} The target object was severely occluded by other objects.
    \item {Multiple Objects:} The candidate region enclosed multiple objects.  % Name: [仮]
\end{itemize}

As shown in Table \ref{tab:error}, the main bottleneck was CE. 
We could reduce the number of cases by using a huge number of source samples or introducing pretrained models \cite{radford2021learning,jia2021scaling} that embed language features and visual features into the same embedding space.

% Qi et al. \cite{qi2020reverie} report that humans achieved an accuracy of 90.76\% for the REC task on REVERIE.
% The goal of the task is to select the target object from all the context objects in the panorama observed from the goal point of the navigation task. Therefore, this task is more complex than the REVERIE-fetch task; human performance on REVERIE-fetch is expected to be higher than 90.76\%.

% \vspace{-1mm}
% \newpage
\section{Conclusions}
In this study, we proposed PCTL, which is a novel transfer learning approach for the multimodal language understanding task. Specifically, we developed Dual ProtoNCE, which is a new contrastive loss generalized for transfer learning.

Our key contributions are as follows:
\begin{itemize}
    \item We introduced transfer learning to the MLU-FI task.
    \item We proposed PCTL, which is a novel transfer learning approach for the multimodal language understanding task.
    \item Within PCTL, we developed Dual ProtoNCE, which is a novel contrastive loss generalized for transfer learning.
    \item PCTL outperformed the baselines in terms of the accuracy of the MLU-FI task on the REVERIE-fetch dataset.
\end{itemize}

In future work, we plan to enrich the source-domain dataset using a simulator and apply the model trained by the PCTL framework to physical robots.

% \appendix
% \input{sectionA}

\section*{ACKNOWLEDGMENT}
This work was partially supported by JSPS KAKENHI Grant Number 20H04269, JST Moonshot, and NEDO.

\bibliographystyle{IEEEtran}
\bibliography{reference}

\end{document}